\documentclass{article}
\usepackage{ijcai22}

\usepackage{times}
\usepackage{soul}
\usepackage{url}
\usepackage[hidelinks]{hyperref}
\usepackage[utf8]{inputenc}
\usepackage[small]{caption}
\usepackage{graphicx}
\usepackage{amsmath}
\usepackage{amsthm}
\usepackage{booktabs}
\usepackage{algorithm}
\usepackage{algorithmic}
\usepackage{multirow}
\usepackage{color}
\usepackage{paralist}
\usepackage{footmisc}

\title{\textsc{CounterGeDi}: A controllable approach to generate\\ polite, detoxified and emotional counterspeech}

\author{
Punyajoy Saha$^1$
\and
Kanishk Singh$^1$\and
Adarsh Kumar$^1$\and
Binny Mathew$^1$ \And
Animesh Mukherjee $^1$
\affiliations
$^1$Indian Institute of Technology, Kharagpur, India\\
\emails
\{punyajoys,kanishksingh,adarshkumar712,binnymathew\}@iitkgp.ac.in,
animeshm@cse.iitkgp.ac.in
}

\begin{document}

\maketitle

\begin{abstract}
  Recently, many studies have tried to create generation models to assist counter speakers by providing counterspeech suggestions for combating the explosive proliferation of online hate. However, since these suggestions are from a vanilla generation model, they might not include the appropriate properties required to counter a particular hate speech instance. 
  In this paper, we propose \textsc{CounterGeDi} - an ensemble of generative discriminators (\textsc{GeDi}) to guide the generation of a DialoGPT model toward more polite, detoxified, and emotionally laden counterspeech. We generate counterspeech using three datasets and observe significant improvement across different attribute scores. The politeness and detoxification scores increased by around 15\% and 6\% respectively, while the emotion in the counterspeech increased by at least 10\% across all the datasets. We also experiment with triple-attribute control and observe significant improvement over single attribute results when combining complementing attributes, e.g., \textit{politeness}, \textit{joyfulness} and \textit{detoxification}. In all these experiments, the relevancy of the generated text does not deteriorate due to the application of these controls.

\end{abstract}

\section{Introduction}

One of the most effective strategies to combat the rising online hate speech is \textit{counterspeech}. It is a direct response to hateful or harmful speech that seeks to undermine it. Several organisations like Facebook\footnote{\url{https://counterspeech.fb.com/en/}} have laid out guidelines for general public on how to counter hateful speech online. While these guidelines might be effective, writing a proper counterspeech is quite challenging~\cite{fumagalli2020counterspeech}. Recently, many non-profit organisations have taken up the task of countering online hate\footnote{https://wecounterhate.com/}. This nichesourcing of counterspeech can be effective but might be a mentally taxing task for the NGO operators given the amount of hate generated each day~\cite{vidgen2019much}. To assist these operators, the scientific community have come up with different human-in-the-loop methods to collect counterspeech data~\cite{fanton2021human} as well as build models to generate counterspeech suggestions~\cite{zhu2021generate}. 

Such generation models aim to reduce the human intervention by helping the counter speakers with suggestions\footnote{We view the counterspeech generating task as delicate task which requires human supervision.}, which they can further post-edit as per requirements. While these recent transformer based generation models can produce relevant outputs, they often fail to produce diverse output~\cite{holtzman2019curious}. Further, we cannot control the generated output for any attribute from the vanilla generation model. However, as pointed out by different authors~\cite{bartlett2015counter}, counterspeech can vary based on the hate speech instance, demography of the hate and counter speakers~\cite{mathew2019thou} etc. Hence, the generation models without any control might produce suggestions that are not suitable for a particular instance.
To counter effectively, the counter speakers should have the control over the generated content so that they can steer the counterspeech toward a desired property or a mixture of different properties. Such a generation model will allow the counter speaker to use different strategies to counter hate~\cite{benesch2016considerations}. In this paper, we propose six different \textsc{GeDi}\footnote{Generative discirimintators.} models and use them in single and multiple attribute setting to guide the generation of a vanilla counterspeech generation model for English language.  We observe that ---
\begin{compactitem}
    \item The single attribute \textsc{GeDi} models improve the controlled attribute significantly for all the attributes. Specifically, there is 15\% increase in politeness and 6\% increase in detoxification. 
    \item Among the \textsc{GeDi} with emotional attributes, \textit{joy} has the highest attribute scores with an improvement of 15\%, 17\% and 88\% in Gab~\cite{qian2019benchmark}, Reddit~\cite{qian2019benchmark}, and CONAN~\cite{chung2019conan} datasets respectively.
    \item While combining multiple attributes --- \textit{politeness}, \textit{detoxification} and \textit{joyfulness} (one of the emotional attribute) for controlling the generation, we observe a further increase in scores compared to the single attribute setup.  
    \item We are able to maintain the relevancy of the generated output measured using \textbf{BLEU-2} score while there is a slight drop in performance in terms of fluency which comes as a cost for the additional control. 
    \item Finally, human judges also marked higher presence of the attributes in the attribute controlled setups compared to the baselines. 
\end{compactitem}
We show examples of emotion controlled counter speech corresponding to a hate speech in Table~\ref{tab:table-example1}. Please refer to Appendix for more examples on single and multi attribute controlled counterspeech. The code is added here\footnote{\url{https://github.com/hate-alert/CounterGEDI}}.
 
\begin{table}[!htpb]
\centering
\scriptsize{
\begin{tabular}{p{1.8cm}p{5.8cm}}\hline
Hate speech              &  \textit{Anyone who isn't a re***ded pickle has privilege, I suppose}.\\\hline
Counterspeech (no control)  &  \textit{The R word is highly offensive, Spread the word to end the word!}\\\hline
Counterspeech (sad)    &  \textit{in the words of my favorite friend this post sounds like such hate speech. it makes you sound uneducated. \textbf{how sad}. do not use internet powers to silence your anger with something so simple or even reasonable!}\\\hline
Counterspeech (joy) &  \textit{we all believe in open mindedness, whether your side of the fence is greater or lesser. \textbf{we hope you have fun!} mwahaha peace out and cool weather. keep hydrated npcs.}\\\hline
\end{tabular}}
\caption{\scriptsize{This table presents an example hate speech and the counterspeech generated using different control attributes which is marked in the (bracket). Please refer to the Appendix for more examples on single and multi attribute controlled counterspeech.}} 
\label{tab:table-example1}
\end{table}

\section{Related works}
With regard to countering hatred, we focus on two research aspects relevant to the present work (i) counterspeech datasets and models for counterspeech generation (ii) controlling methods for generation models particularly in NLP. 

\noindent\textbf{Counterspeech generation strategies}: An effective strategy to mitigate hate speech is counterspeech as it does not violate freedom of expression~\cite{benesch2016considerations}. While the idea of countering some hateful messages is not new
, the research community has recently started taking a massive interest in understanding counterspeech practices and its effectiveness while mitigating hate speech~\cite{mathew2019thou}. 
Recently, \citeauthor{tekirouglu2020generating} proposed novel techniques to generate counterspeech using a GPT-2 model with post-facto editing by experts or annotator groups.  
One of the recent generation methods uses a three stage pipeline -- \textit{Generate, Prune} and \textit{Select} (GPS) to generate diverse and relevant counterspeech output~\cite{zhu2021generate}. 
The primary challenge in counterspeech generation research is understanding if the counterspeech produced is effective. 
One study\cite{bartlett2015counter} found that effectiveness of the counterspeech further depends on the tone of the counterspeech. In specific, sentimental or casual tone received 83\% more responses~\cite{frenett2015one}. In addition to that, \citeauthor{mathew2019thou} found that different communities find different types of counterspeech effective. 
In this work, we present a novel approach to guide the counterspeech toward one or more desired properties. This is done by building a controllable counterspeech generation pipeline based on \textsc{GeDi}.

\noindent\textbf{Controllable text generation}: Controllable text generation is the task of generating natural sentences whose attributes can be controlled. The previous approaches rely on reinforcement learning or training conditional generative models~\cite{prabhumoye2020exploring}.   
One of the earliest line of work focused on controlling a desired attribute by side constraints~\cite{sennrich-etal-2016-controlling} and back propagating gradients~\cite{dathathri2019plug}. 
One of the variations - \textsc{DExperts}~\cite{liu2021dexperts} uses language models for both positive and negative classes. While the other variation - \textsc{GeDi}~\cite{krause2020gedi} uses class conditioned language models for positive and negative classes. The final output tokens in both these methods are generated based on an equation which utilises the contrast between the positive and the negative class. 

We, in this work, use this \textsc{GeDi} model and apply it to the domain of counterspeech generation. We train different \textsc{GeDi} models to control attributes like \textit{politeness}, \textit{emotions} and \textit{detoxification} of counterspeech.

\section{Models}

\noindent\textbf{DialoGPT}: We used a variant of the GPT model - DialoGPT~\cite{zhang2020dialogpt} which was trained on a large corpus consisting of English Reddit dialogues. The corpus consist of 147 million instances of dialogues, collected over a period of 12 years. Unlike GPT-2, this model should generate better dialouge like responses to any given prompt. In this model, along with ground truth response $T=x_1,...,x_n$ we also have a dialogue utterance history $S$. The model aims at maximising $p(T|S) = p(x_1|S)\prod^n_{i=2} p(x_i|S,...,x_{i-1})$. For our experiment, we used DialoGPTm - a 24 layer, 345 million weight parameters transformer model\footnote{https://huggingface.co/microsoft/DialoGPT-medium} and finetune over a particular dataset having hate and counterspeech pairs.

\noindent\textbf{Generate Prune Select (GPS)}: One of the recent counterspeech generation model is \textit{Generate, Prune, Select} (\textit{GPS}) -- a three stage pipelined approach~\cite{zhu2021generate}. At first, the \textit{generation part} generates a large number of diverse response candidates using a generative model based on RNN based autoencoder. Second, the \textit{pruning part} prunes the ungrammatical candidates from the candidate pool. This is done using a classifier trained on linguistic acceptability classifier. Finally, the \textit{response-selection part} selects appropriate responses based on the hate speech instance. We use the similarity based method \textit{USE-LARGE-SIM}~\cite{zhu2021generate}.

\noindent\textbf{\textsc{GeDi}}: For controlling generated counterspeech,  we use a recent method Generative Discriminators (\textsc{GeDi})~\cite{krause2020gedi}, where the authors present a decoding time algorithm to control the output from the generation model. \textsc{GeDi} assumes we have class conditioned language model (CC-LM) with a desired control code $c$ and an undesired control code $\bar{c}$. For our case, we fix the control code $c$ as `true' and $\bar{c}$ as `false'. For each dataset, the attribute mentioned in the $+ve$ column in Table \ref{tab:attribute_dataset} is considered as desired, while $-ve$ column is used as undesired attribute.

The authors use the contrast between $P_{\theta}(x_{1:t}|c)$ and $P_{\theta}(x_{1:t}|\bar{c})$ to guide sampling from an LM that gives $P_{LM}(x_{1:T})$. The probability that the next token $x_t$ belongs to desired class is calculated using this contrast. \if{0}This is calculated using equation~\ref{eq:1} where $\alpha$ is a learnable scale parameter.

\begin{equation}
\scriptsize
    \label{eq:1}
    P_{\theta}\left(c \mid x_{1: t}\right)=\frac{P(c) P_{\theta}\left(x_{1: t} \mid c\right)^{\alpha / t}}{\sum_{c^{\prime} \in\{c, \bar{c}\}} P\left(c^{\prime}\right) P_{\theta}\left(x_{1: t} \mid c^{\prime}\right)^{\alpha / t}}
\end{equation}

In order to train the \textsc{GeDi} model, the authors combine the generative language modeling loss $\mathcal{L}_g$ (ref equation \ref{eq:lg}) as shown in equation~\ref{eq:2} with a discriminative loss $\mathcal{L}_d$. The equation \ref{eq:3}  shows the final loss $\mathcal{L}_{gd}$ where $\lambda$ is a learnable parameter.

\begin{equation}
\scriptsize
    \label{eq:lg}
    \mathcal{L}_{g}=-\frac{1}{N} \sum_{i=1}^{N} \frac{1}{T_{i}} \sum_{t=1}^{T_{i}} \log P_{\theta}\left(x_{t}^{(i)} \mid x_{<t}^{(i)}, c^{(i)}\right)
\end{equation}

\begin{equation}
\scriptsize
    \label{eq:2}
    \mathcal{L}_{d}=-\frac{1}{N} \sum_{i=1}^{N} \log P_{\theta}\left(c^{(i)} \mid x_{1: T_{i}}^{(i)}\right)
\end{equation}

\begin{equation}
\scriptsize
    \label{eq:3}
    \mathcal{L}_{g d}=\lambda \mathcal{L}_{g}+(1-\lambda) \mathcal{L}_{d}
\end{equation}\fi

For controlled generation, the authors propose a simple method to guide the model toward the target class which is represented using the heuristic equation \ref{eq:4} where $\omega$ is controllable parameter. In order to control multiple attributes, we extend the heuristic as represented in equation \ref{eq:5} where $\omega_{i}$ is a controllable parameter to bias the generation toward class $c_{i}$ for \textsc{GeDi} trained on the $i^\textrm{th}$ attribute. 

\begin{equation}
    \scriptsize
    \label{eq:4}
    P_{w}\left(x_{t} \mid x_{<t}, c\right) \propto P_{L M}\left(x_{t} \mid x_{<t}\right) P_{\theta}\left(c \mid x_{t}, x_{<t}\right)^{\omega}
\end{equation}

\begin{equation}
    \label{eq:5}
    \scriptsize
    P_{w}\left(x_{t} \mid x_{<t}, c_{1}, .. c_{n}\right) \propto P_{L M}\left(x_{t} \mid x_{<t}\right)\prod_{i=1,..,n} P_{\theta}\left(c_{i} \mid x_{t}, x_{<t}\right)^{\omega_{i}}
\end{equation}

\section{Datasets}

\noindent\textbf{Counterspeech datasets}: In order to evaluate our approach we use three public datasets which contain hate speech and its corresponding counterspeech. The details of these datasets are noted in Table \ref{tab:cs_dataset}. Reddit and Gab datasets contain $5,257$ and $14,614$ hate speech instances respectively~\cite{qian2019benchmark}. We use the English part of the CONAN dataset~\cite{chung2019conan} which contains $408$ hate speech instances.  The counterspeech in Gab and Reddit datasets were written by AMT workers, whereas for CONAN the counterspeech was written by expert NGO operators.

We further made hate speech and counterspeech pairs from these datasets such that each hate speech was associated with one counterspeech. Finally, we ended up with 3,864, 14,223, 41,580 datapoints for CONAN, Reddit and Gab respectively. We split each dataset randomly into train, validation, test set with 80\% for training, 10\% each for validation and testing.
\begin{table}[!htpb]
\scriptsize
\centering
\begin{tabular}{|l|l|l|l|l|}
\hline
\textbf{Dataset} & \textbf{Source-H} & \textbf{Source-C} & \textbf{Hate instances} & \textbf{Total pairs} \\ \hline
CONAN       & synthetic     & expert     &   408     &   3,864      \\ 
Reddit      & reddit        & crowd     &   5,257    &   14,223     \\ 
Gab        & gab           & crowd      &   14,614    &   41,580     \\ \hline
\end{tabular}
\caption{\scriptsize{This table presents the source of hate speech (Source-H), source of counterspeech (Source-C), hate speech instances and the total pairs for each of the CONAN, Reddit and Gab dataset.}}
\label{tab:cs_dataset}
\end{table}

\noindent\textbf{Attribute datasets}: We control several attributes in the generated counterspeech. We selected these attributes following the recommended strategies for counterspeech~\cite{benesch2016considerations} and properties of responses in human conversation~\cite{10.1145/3290605.3300705}.

\noindent\textit{Politeness}: One of the properties of counterspeech as suggested by \cite{benesch2016considerations} is empathy. As a first step in that direction we tried to make the generated counterspeech more polite. We used the dataset of 1.39 million posts released by~\cite{madaan2020politeness} labelled into nine politeness classes (P1-P9). As recommended by the authors, we considered P9 as the polite part and others (P1-P8) as non-polite.

\noindent\textit{Detoxification}: \cite{benesch2016considerations} also noted several strategies which are discouraged while writing counterspeech. One of these discouraged strategies are hostile or aggressive behaviour. 
To detox any hostile counterspeech generated by the generation model, we use the a popular Kaggle dataset\footnote{https://www.kaggle.com/c/jigsaw-toxic-comment-classification-challenge/data} which contains text samples having `toxic' and `non-toxic' labels. 
We stratified-split the released training dataset randomly into 90\% training and 10\% validation sets. The test set is already released separately with the dataset. 
We trained a \textsc{GeDi} model considering toxic as the positive label and non-toxic as the negative label. While generating using \textsc{GeDi}, we guide the generation toward the negative class (non-toxic).

\noindent\textit{Emotion:} Another important aspect of conversation is communicating different emotions. A study~\cite{prendinger2005empathic} found that systems expressing emotions are more capable of providing user satisfaction. In case of counterspeech, emotions might enhance the effect of the generated counterspeech~\cite{benesch2016considerations}. For example, `sadness' as an emotion can be added when the counter speakers affiliate themselves with the target group. Similarly, `joy' can used to convey positivity in the counterspeech.

In order to control the emotion while generating a counterspeech, we used a large dataset~\cite{saravia2018carer} of 416,809 datapoints comprising posts having seven emotions -- `sadness', `joy', `fear', `anger', `surprise', and `love'. For this paper, we did not consider - `love' and `surprise' emotions as these had less than 10\% posts in the dataset. We stratified-split each dataset randomly into training, validation, and test set with 80\% for training, and 10\% for both validation and testing. We consider each emotion as a separate attribute and trained a \textsc{GeDi} model for that emotion by considering it as positive label and other emotions as negative labels. For our experiments, we primarily focus on guiding the models toward the positive class.

A summary statistic of the attribute dataset for each of the task considered is noted in Table \ref{tab:attribute_dataset}.

\begin{table}[!htpb]
\scriptsize
\centering
\begin{tabular}{|l|c|c|c|c|c|}
\hline
\textbf{Dataset}    & \textbf{+ve} & \textbf{-ve} & \textbf{T\textsubscript{r} (\%+ve)} & \textbf{V (\%+ve)} & \textbf{T\textsubscript{e} (\%+ve)} \\ \hline
Polite & p     &   n-p  & 1.12M (20\%) & 137k (20\%) & 137k (20\%)\\ \hline
Toxic   & t        &   n-t   & 143k (10\%) & 16k (10\%) & 153k (4\%)\\ \hline
\multirow{4}{*}{Emotion}    & j           &   o      & 333k (34\%) & 42k (34\%) & 42k (34\%)\\
    & f           &   o      & 333k (11\%) & 42k (11\%) & 42k (11\%)\\ 
    & s           &   o      & 333k (29\%) & 42k (29\%) & 42k (29\%)\\ 
    & a           &   o      & 333k (14\%) & 42k (14\%) & 42k (14\%)\\ \hline
\end{tabular}
\caption{\scriptsize{This table shows the attribute datasets, positive and negative classes and data present in train, validation and test part for each. T\textsubscript{r}: Train, V: Validation, T\textsubscript{e}: Test, p: polite, n-p: non-polite, t: toxic, n-t: non-toxic, s: sadness, j: joy, a: anger, f: fear, o: others. The \% associated with the  T\textsubscript{r}, V and  T\textsubscript{e} are the \% of positive labels.}}
\label{tab:attribute_dataset}
\end{table}

\section{Experimental setup}
\noindent\textbf{Counterspeech generation models}: The DialoGPTm model for each counterspeech model has six initial layers fixed due to resource constraints. The model were trained till 10 epochs with batch size as 8. We saved the final model at the epoch having the best language modelling loss for the validation dataset. We used a maximum length of 256 tokens for the DialoGPTm model\footnote{1\% datapoints have more than 256 tokens.}. The learning rate is fixed at $5e^{-6}$ for training the model.

\noindent\textbf{\textsc{GeDi} models}: We train the GPT-2 model as the \textsc{GeDi} models based on the training setting specified in the original paper~\cite{krause2020gedi}. For each model we fix the batch size at 8  and train the models for 5 epochs. The $\lambda$ weight in the loss equation is fixed at 0.8 to maximise generation quality for the \textsc{GeDi} model. We used a maximum length of 128 tokens for the GPT-2 model.  The learning rate is fixed at $2e^{-5}$ for training the model following the recommendations by \cite{krause2020gedi}. 

\begin{figure}
    \centering
    \includegraphics[width=0.4\textwidth]{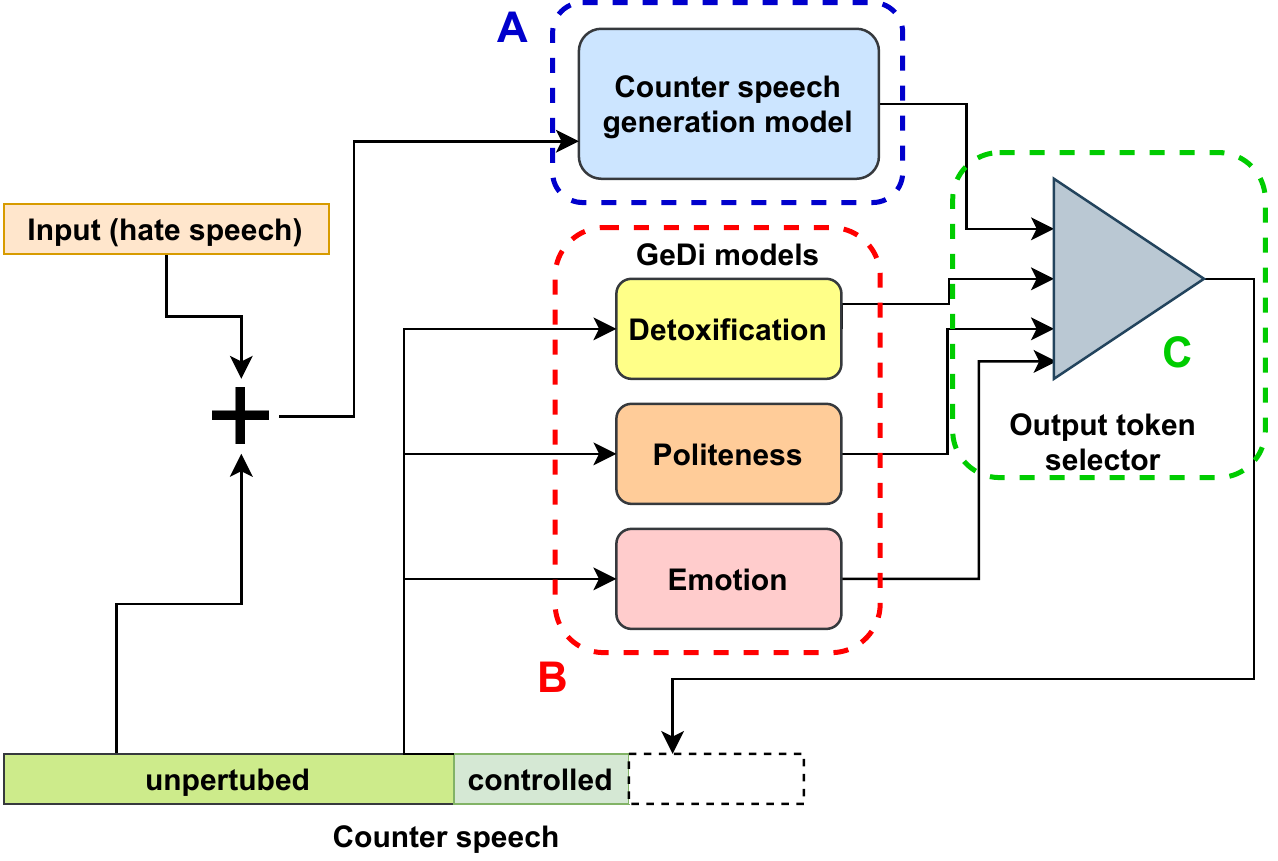}
    \caption{\scriptsize{The figure shows how the overall setup of the pipeline. The counterspeech generation model is produces a probability distribution for the next possible token using hate speech+ counterspeech generated till now. The \textsc{GeDi} models individually take in the currently generated counter speech and produces a probability distribution for the next possible token based on the desired attributes. The output token selector selects the next output token. The unperturbed part in the counterspeech is created without any control, to provide the initial prompt to the \textsc{GeDi} model.}}
    \label{fig:pipeline}
\end{figure}

\noindent\textbf{Final pipeline}: Our final pipeline comprises three parts as shown in Figure \ref{fig:pipeline}. The \textbf{part A} represents the vanilla counterspeech generation model trained on one of the three counterspeech datasets. Similar to an auto-regressive setup, it takes in the hate speech with the currently generated counterspeech (empty at the initial step) and produces next token probablities for the production of the next token. The \textbf{part B} consists of the single or multiple \textsc{GeDi} models. Each \textsc{GeDi} model controls an attribute out of the total six. It takes as input the currently generated counterspeech and produces token probabilities based on the desired attribute. We initially allow the counterspeech generation models to generate 10 tokens without any control to provide the initial prompt to the \textsc{GeDi} model. Finally, \textbf{part C} selects the next token based on the token probabilities from different models, i.e., the counterspeech generation model and the \textsc{GeDi} models following equation \ref{eq:5}. For each \textsc{GeDi} model, we primarily control the weight ($\omega$) as mentioned in equation \ref{eq:4}, while other parameters are kept same as the paper~\cite{krause2020gedi}. For single attribute, we fix the weight at 1 to give equal importance to the counterspeech generation as well as the control attributes. For two attribute control, we set weights at 0.5 for both the attributes. For three attributes control, which comprises detoxification, politeness and an emotion, we set 0.3 for politeness \& detoxification each and 0.4 for the emotion. We also use nucleus sampling as a decoding strategy~\cite{holtzman2019curious}. Please check Appendix for more details.

\section{Evaluation}
We consider several metrics to evaluate our whole pipeline of controlled counterspeech generation. The \textit{generation metrics} measure the generation capability of the DialoGPTm and \textsc{GeDi} models. The \textit{classification metrics} are mainly to evaluate the \textsc{GeDi} model on the attribute datasets. Finally, we measure the amount of control in the generated counterspeech using external classifiers which we refer to as \textit{controller metrics}. We generate 5 samples for every hate speech instance with DialoGPTm. The GPS framework automatically selects the best response based on the heuristic, hence we keep one sample for every hate speech instance.

\noindent\textbf{Generation metrics}: To measure the generation quality, we use different standard metrics. We use \textit{BLEU-2}\footnote{converted to a scale of 0-100 from 0-1} and \textit{METEOR}~\cite{sai2020survey} to measure how similar the generated counterspeech are to the ground truth counterspeech. We also measure if the generation model generates a diverse and novel counterspeech using metrics from previous research~\cite{wang2018sentigan}. To measure fluency, we use a classifier of linguistic acceptability trained on the COLA dataset~\cite{warstadt2019linguistic}

\noindent\textbf{\textsc{GeDi} metrics}: For classification, we report \textit{accuracy}, \textit{macro F1-score}, and \textit{AUROC} score for each \textsc{GeDi} model's performance on a test dataset of a particular attribute. We also report the generation performance using the perplexity~\cite{zhang2020dialogpt}.

\noindent\textbf{Controller metrics}: In order to evaluate the ability of the \textsc{GeDi} controller to control the attribute, we used third-party classifiers for each attribute. For politeness, we trained a bert-base-uncased model for politeness level detection on a scale of 0 to 7\footnote{\footnotesize https://github.com/AlafateABULIMITI/politeness-detection}. For measuring emotion in the generated text, we used the Ekman version of the GoEmotions models\footnote{\footnotesize https://huggingface.co/monologg/bert-base-cased-goemotions-ekman}. For each post, it returns a confidence score between 0-1 for anger, disgust, fear, joy, sadness, surprise + neutral.  We report the confidence score for a particular emotion as a measure of that emotion in a given post. Finally, to measure toxicity we used the HateXplain model~\cite{mathew2020hatexplain} trained on two classes -- toxic and non-toxic\footnote{\footnotesize https://huggingface.co/Hate-speech-CNERG/bert-base-uncased-hatexplain-rationale-two}. We report the confidence between 0-1 for the non-toxic class.

\section{Results}

\noindent\textbf{Generation results}: We compare DialoGPTm model with the GPS in Table \ref{tab:results-accurate}. We find ~\textbf{BLEU-2} scores are better for the GPS model while the \textbf{METEOR} scores are better for DialoGPT model for all three datasets. DialoGPTm is also better in terms of novelty and diversity for all the three datasets. The fluency metric \textbf{COLA} is better for GPS for all the three datasets, since it straightforwardly prunes grammatically incorrect samples. Since DialoGPTm presents a competitive performance compared to the state-of-the-art model, we therefore use DialoGPTm for the rest of the experiments.

\begin{table}[!htpb]
\centering
\scriptsize
\begin{tabular}{|c|c|c|c|c|c|}
\hline
Model & B2 ($\uparrow$) & COLA ($\uparrow$) & M ($\uparrow$) & N ($\uparrow$) & D ($\uparrow$)\\
\hline
\multicolumn{6}{|c|}{\textbf{CONAN}}                                   \\ \hline

GPS & \textbf{41.5} & \textbf{0.82} & 0.14  & 0.18 & 0.60  \\
DialoGPTm & 12.7& 0.78& \textbf{0.18} &\textbf{ 0.84} & \textbf{0.80} \\\hline

\multicolumn{6}{|c|}{\textbf{Reddit}}                                   \\ \hline

 GPS & \textbf{14.1} & \textbf{0.82} & 0.11 & 0.30 & 0.47  \\
 DialoGPTm & 6.9 & 0.75 & \textbf{0.17} & \textbf{0.82} & \textbf{0.74} \\\hline

\multicolumn{6}{|c|}{\textbf{Gab}}                                   \\ \hline

GPS & \textbf{13.9 }& \textbf{0.82} & 0.12 & 0.15  & 0.41 \\
DialoGPTm & 7.7 & 0.80 & \textbf{0.17} & \textbf{0.80} & \textbf{0.72} \\
\hline
\end{tabular}
\caption{\scriptsize{Evaluation results for the three datasets. We report BLEU-2 (B2), COLA, METEOR (M), novelty (N) and diversity (D) to compare the two baselines: generate-prune-select (GPS) framework and DialoGPTm. For all metrics, higher is better and \textbf{bold} denotes the best scores.}}
\label{tab:results-accurate}
\end{table}

\noindent\textbf{\textsc{GeDi} metrics}: As reported in Table \ref{tab:attribute-performance}, we find that F1-score and AUCROC scores for politeness and all the four emotions are above 0.9. This highlights that even with $0.2$ as the weight for the discriminator we are able to get good scores on classification. The perplexity scores for all the test datasets are also around $3.5$\footnote{For reference, perplexity for pretraining GPT-2 comes around 10 after 10K steps (https://tinyurl.com/3vwrvscd).}. \textsc{GeDi} model for toxicity has lower scores than the other attribute tasks. The F1-score for toxicity detection is $\sim0.6$ and AUCROC is $\sim 0.83$. The perplexity is also higher at around $4.5$ for the toxicity dataset. This highlights the difficulty of the task of detecting toxicity.
\begin{table}[h!]
\scriptsize
\centering
\begin{tabular}{cc|c|c|c|c}
\hline
\textbf{Dataset} & \textbf{Positive} & \textbf{F1} ($\uparrow$) & \textbf{Acc} ($\uparrow$) & \textbf{AUC}($\uparrow$) & \textbf{Perplexity} ($\downarrow$)  \\ \hline
Toxicity & toxic  &   0.60 & 0.85  & 0.84 & 4.428\\ \hline
Politeness & polite  & 0.93 & 0.96   & 0.93 & 3.476 \\ \hline
Emotion & joy  & 0.96 & 0.96 & 0.97  & 3.546 \\ %\hline
Emotion & sadness  & 0.98 & 0.98 & 0.99  & 3.543 \\ %\hline
Emotion & fear  &  0.94 & 0.97  & 0.98 & 3.774\\ %\hline
Emotion & anger  & 0.96 & 0.98  & 0.99 & 3.560 \\ \hline
\end{tabular}
\caption{\scriptsize{\textsc{GeDi} generation and classification performance on test set of attribute datasets. Generation is evaluated using the Perplexity whereas classification performance is measured using F1-score (F1), Accuracy (Acc) and AUCROC (AUC). For all the metrics except perplexity, higher is better.}}
\label{tab:attribute-performance}
\end{table}

\noindent\textbf{Single-attribute control}: In Table \ref{tab:control-single}, we report the amount of different attributes present in the generated counterspeech for each dataset and for each model. When we compare GPS and DialoGPTm, we find that except anger emotion, all other scores are significantly higher for DialoGPTm. Second, using control for a particular attribute significantly improves the presence of that attribute (p-value $<$0.001). For instance, in Table \ref{tab:control-single}, the politeness score increases from 3.91 to 4.54, from 5.24 to 6.05 and 5.14 to 6.11 for CONAN, Reddit and Gab respectively when the DialoGPTm model is controlled for politeness. This is true for all attributes barring the `anger' emotion. Politeness and detoxification score increased by 15-18\% and 6-8\% respectively across all the datasets. For the emotion attributes, `joy' has the highest scores among all for both controlled and uncontrolled attribute. We see an overall increase in `joy' of around 17\% for Gab, 14\% for Reddit and 88\% for CONAN. Counter responses in CONAN datasets are mostly devoid of any emotions hence bringing a change in them is much easier than the Reddit/Gab datasets which are higher in terms of the joy attribute. We reach closer to GPS baseline for anger emotion while controlling anger emotion and increase the score by 54\%, 55\% and 16\% for Reddit, Gab and CONAN, respectively. While the increase for other emotions -- `sadness' and `fear' increased significantly, the overall scores for them remain low.

\begin{table}[h!]
\centering
\scriptsize
\begin{tabular}{|c|c|c|c|c|c|c|}
\hline
\textbf{Model} & \textbf{D} ($\uparrow$)  & \textbf{P} ($\uparrow$) & \textbf{J} ($\uparrow$) & \textbf{A} ($\uparrow$) & \textbf{S} ($\uparrow$) & \textbf{F} ($\uparrow$)\\
\hline
\multicolumn{7}{|c|}{\textbf{CONAN}}                                   \\ \hline
 GPS & \textbf{0.68} & 2.01 & 0.16 & \textbf{0.12} & 0.03 & 0.01 \\
 DialoGPTm & 0.64 & 3.91 & 0.18 & 0.09 & 0.04 & 0.01\\ 
 DialoGPTm-c & \textbf{0.68} & \textbf{4.54} & \textbf{0.34} & 0.11 & \textbf{0.08} & \textbf{0.05}\\\hline 
\multicolumn{7}{|c|}{\textbf{Reddit}} \\\hline
 GPS & 0.82 & 1.62 & 0.23 &\textbf{0.32} & 0.04 & 0.01 \\
 DialoGPTm & 0.82 & 5.24 & 0.63 & 0.17 & 0.06 & 0.00\\
 DialoGPTm-c & \textbf{0.87} & \textbf{6.05} & \textbf{0.72} & 0.27 & \textbf{0.10}& \textbf{ 0.02}\\ \hline
 \multicolumn{7}{|c|}{\textbf{Gab}} \\\hline
GPS & 0.79 & 1.46 & 0.22 & \textbf{0.28} & 0.04 & 0.01 \\
DialoGPTm & 0.81 & 5.14 & 0.66 & 0.17 & 0.05 & 0.00\\
DialoGPTm-c & \textbf{0.85} & \textbf{6.11} & \textbf{0.77} & 0.26 & \textbf{0.10} & \textbf{0.02}\\
\hline
\end{tabular}
\caption{\scriptsize{Performance of single attribute setups with the vanilla baseline generate-prune-select (GPS) and  DialoGPTm models. Each column name represents the attribute being measured. The attributes measured are politeness (P), detoxification (D), sadness (S), joy (J), anger (A) and fear (F). Politeness (P) is measured in a scale of 0-7 whereas others are measured in the scale $[0,1]$. For the last row - controlled DialoGPTm (DialoGPTm-c) the column name also represents the attribute getting controlled. For all the metrics, higher is better and \textbf{bold} denotes the best scores.}}
\label{tab:control-single}
\end{table}

%%%%%%%%%%%%%%%%%%%%%% BLEU and COLA for single attributes %%%%
\begin{table}[h!]
\centering
\scriptsize
\begin{tabular}{|c|c|c|c|c|c|c|}
\hline
\textbf{Scores} & \textbf{Detox}  & \textbf{Polite}  & \textbf{Joy} & \textbf{Anger} & \textbf{Sadness} & \textbf{Fear}\\
\hline
\multicolumn{7}{|c|}{\textbf{CONAN}}                                   \\ \hline
 BLEU-2 & \textbf{13.8} & 12.1 & 12.2 & 11.6 & 12.0 & 12.8 \\
 COLA & \textbf{0.83} & 0.72 & 0.72 & 0.74 & 0.76 & \textbf{0.72}\\\hline 
\multicolumn{7}{|c|}{\textbf{Reddit}} \\\hline
 BLEU-2 & \textbf{8.1} & 7.8 & 7.7 & 7.8 & 7.5 & 7.3 \\
 COLA & 0.72 & 0.77 & 0.70 & 0.72 & \textbf{0.81}&  0.70\\ \hline
 \multicolumn{7}{|c|}{\textbf{Gab}} \\\hline
BLEU-2 & \textbf{8.7} & 8.3 & 8.5 & 8.3 & 8.2 & 8.3 \\
COLA & \textbf{0.85} & 0.82 & 0.76 & 0.76 & 0.80 & 0.78\\
\hline
\end{tabular}
\caption{\scriptsize{BLEU-2 and COLA performance for single attribute setups for DialoGPTm-c model. Each column name represents the individual attribute model namely politeness (P), detoxification (D), sadness (S), joy (J), anger (A) and fear (F). \textbf{Bold} denotes the best scores across the row.}}
\label{tab:control-single-relevance}
\end{table}

\noindent\textbf{Multi-attribute control}: We also generate counterspeech with the DialoGPTm with mutli-attribute control. We keep politeness, detoxification and one of the emotion\footnote{One among `joy', `anger', `fear' and `sad'.} as control attributes. This gives us four variations for each dataset. We then measure the individual attribute scores for each of these three attribute and report the results in Table \ref{tab:multi-attribute}. For detoxification scores, the setup - $joy+polite+detox$ outperforms other setups across all the experiment. This setup even outperforms the single-attribute detoxification setup by 8\%, 2\% and 2\% for CONAN, Reddit and Gab, respectively. For politeness score, the best performance occurs for $joy+polite+detox$ setup for CONAN and Reddit dataset, while the setup - $fear+polite+detox$ performs better in case of the Gab dataset. Compared to single attribute setup for politeness, the politeness scores drop across all the multi-attribute setups. Among the emotions, the attribute score for `joy' in a multi-attribute setting outperforms the single attribute setting by 44\%, 13\% and 10\% for CONAN, Reddit and Gab.  For `anger', the scores in multi-attribute setting decrease around 25-30\% when compared to the single attribute setting. For other attributes like `sadness' and `fear', the multi-attribute results are below 0.1, similar to the single attribute results. Please also see Appendix for attribute ablation performances.

\noindent\textbf{Quality of controlled generation:} In the previous section, we observed that we were able to control attributes in generated outputs in single and multi-attribute setups. While this is encouraging, it is important to understand if the controlled text are losing the central theme of remaining a counterspeech and are still fluent. For the former, we measure the \textbf{BLEU-2} metric and for the latter we use the \textbf{COLA} metric. 

According to Table~\ref{tab:control-single-relevance}, we find that relevance of the output (measured using BLEU-2) does not change much across different attributes for the single attribute setups. For some of the attributes like detoxification, the BLEU-2 scores even outperform the DialoGPTm model (without control) for all the datasets as noted in column B2 in Table \ref{tab:results-accurate}. For Reddit and Gab, there is a further improvement of 1-2 points in the \textbf{BLEU-2} metric for other attributes also as compared to the vanilla DialogGPTm model (in column B2 in Table \ref{tab:results-accurate}). This shows that the controls do not affect the overall relevance of the generated counterspeech. In fact, the relevance improves in many cases. In terms of fluency, we see a slight drop which comes as a cost for controlling different attributes except few cases (comparing column COLA in Table~\ref{tab:results-accurate} and Table \ref{tab:control-single-relevance}). This might be due to the fact that GEDI model is not geared toward maintaining the fluency of the models. The observation holds for the multi attribute setup as well (comparing columns B2 and COLA in Table~\ref{tab:results-accurate} and  Table~\ref{tab:multi-attribute}).

Overall, we observe that it is possible to control the attributes in the generated outputs using the single attributes. Our experiments with multi-attributes further reveals that there are certain complementing attributes for e.g $joy+polite+detox$  which can be used to further increase the single-attributes setups. For other setups, the attribute scores drops below the single attribute setups. Another promising observation is that the control of attribute does not harm the relevance of the generated output as they still remain close to the ground truth. Since GEDI is not geared toward improving the fluency, we see a slight drop in the fluency of the generated outputs. An interesting research direction would be to look into improving attribute and fluency scores while using multi-attribute setups. We have added examples of single and multi-attribute setup in the Appendix. 

\begin{table}[h!]
\scriptsize
\centering
\begin{tabular}{|p{14mm}|c|c|c|p{6mm}|c|}
\hline
\textbf{Attributes}          & \textbf{Detox}($\uparrow$)         & \textbf{Polite}($\uparrow$)        & \textbf{Emotion}($\uparrow$) & \textbf{B2}($\uparrow$) &
\textbf{COLA}($\uparrow$) \\ \hline
\multicolumn{6}{|c|}{\textbf{CONAN}}                                   \\ \hline
Joy(J)+P+D    &   \textbf{0.74}     &  \textbf{4.13}               & 0.49 (J) & 13.4 & \textbf{0.79}      \\ \hline
Anger(A)+P+D  &    0.67             &  3.06 & 0.08 (A)    & 12.6 & 0.68  \\ \hline
Sad(S)+P+D   &   \underline{0.70}   &  3.56                        & 0.07 (S)  & 13.2 & 0.74     \\ \hline
Fear(F)+P+D  &    \underline{0.70}  &  \underline{4.00}            & 0.06 (F)  & \textbf{13.6} & 0.75     \\ \hline

\multicolumn{6}{|c|}{\textbf{Reddit}}                                   \\ \hline
Joy+P+D    &  \textbf{0.89}     & \textbf{5.79} & 0.82 (J) & 8.3 & \textbf{0.81 }    \\ \hline
Anger+P+D  &  0.85              & \underline{4.24}             & 0.19 (A) & \textbf{8.3 } & 0.72     \\ \hline
Sad+P+D   &   \underline{0.87}  & 3.56             & 0.09 (S)  & 8.2 & 0.79     \\ \hline
Fear+P+D  &  \underline{0.87}   & 4.00             & 0.01 (F) & 7.8 & 0.79      \\ \hline
\multicolumn{6}{|c|}{\textbf{Gab}}                                   \\ \hline
Joy+P+D    & \textbf{0.87}        & \underline{5.68}             &  0.85 (J) & \textbf{8.8} & \textbf{0.85}\\ \hline
Anger+P+D  & 0.83      & 4.11             &  0.19 (A) & 8.5 & 0.75 \\ \hline
Sad+P+D    & 0.85     & 4.70             & 0.09 (S) & \textbf{8.8} & 0.84\\ \hline
Fear+P+D   & \underline{0.86}      & \textbf{5.82} & 0.01 (F) &  \textbf{8.8} & 0.83\\ \hline
\end{tabular}
\caption{\scriptsize{Results of controlling three attributes -- politeness, detoxification and one of the emotions in a multi-attribute setting. The columns represent the amount of the attribute present for each setup. The column -- \textit{emotion} represents the score of the emotion shown in the parenthesis that is being controlled for that instance. BLEU(B2) and COLA were also reported for different setups. For all metrics, higher is better and \textbf{bold} denotes the best scores.}}
\label{tab:multi-attribute}
\end{table}

\if{0}
\begin{table}[h!]
\centering
\begin{tabular}{|c|c|c|c|}
\hline
\textbf{Attributes}          & \textbf{Detox}($\uparrow$)          & \textbf{Polite}($\uparrow$)     & \textbf{Emotion}($\uparrow$) \\ \hline
\multicolumn{4}{|c|}{\textbf{CONAN}}                                   \\ \hline
Joy(J)+Polite+Detox    &    0.73             &  3.44               & 0.37 (J)        \\ \hline
Anger(A)+Polite+Detox  &    0.68             &  2.79               & 0.05 (A)       \\ \hline
Sad(S)+Polite+Detox   &     0.69             &  3.20               & 0.03 (S)        \\ \hline
Fear(F)+Polite+Detox  &     0.70             &  3.30               & 0.01 (F)        \\ \hline

\multicolumn{4}{|c|}{\textbf{Reddit}}                                   \\ \hline
Joy+Polite+Detox    &  0.87               & 5.12                & 0.76 (J)       \\ \hline
Anger+Polite+Detox  &  0.82               & 3.46                & 0.09 (A)    \\ \hline
Sad+Polite+Detox    &  0.84               & 3.96                & 0.05 (S)       \\ \hline
Fear+Polite+Detox   &  0.86               & 3.34                & 0.01 (F)       \\ \hline
\multicolumn{4}{|c|}{\textbf{Gab}}                                   \\ \hline
Joy+Polite+Detox    & 0.85                & 5.09                & 0.82 (J) \\ \hline
Anger+Polite+Detox  & 0.80                & 3.41                & 0.08 (A)  \\ \hline
Sad+Polite+Detox    & 0.82                & 4.19                & 0.04 (S) \\ \hline
Fear+Polite+Detox   & 0.85                & 4.69                & 0.00 (F)\\ \hline
\end{tabular}
\caption{\footnotesize{Results of the ablation study. In each of these setups, we remove one of the attribute and re-estimate that attribute's score. The last column -- \textit{emotion} represents the score of the emotion that is being controlled for that instance. For all metrics, higher is better and \textbf{bold} denotes the best scores.}}
\label{tab:multi-attribute-ablation}
\end{table}
\fi

\noindent\textbf{Human evaluation:} In order to understand, if the improvement in the attribute scores across (while controlling different attributes) would be visible to the moderators, we perform a human evaluation on the generated counterspeech. In this experiment, an annotator is shown three sentences - one generated from the GPS pipeline, another generated using DialoGPTm model and finally, one generated using the DialoGPTm model where some attribute $x$ was getting controlled. We hide the type of model from which the post was generated and further shuffle the posts to remove any ordering bias. Next, the annotator was asked to mark the amount of the attribute $x$ in the given three posts on a scale of 0-5 where 0 presents the absence of the attribute while 5 corresponds to the highest presence of that attribute. Five annotators participated in the annotation with each post getting marked by two annotators. The annotators annotated 20 randomly selected triplets per dataset for each attribute. We do not include the detoxification attribute for these experiments as there is very little difference in detoxification scores when comparing the baseline and the controlled setups. For more details about the annotations, please refer to the Appendix.

We observe an improvement in most of the attribute scores for the controlled model over the two baselines. Three cases where the improvement is not present is while controlling `joy' and 'sad' for the CONAN dataset and controlling `fear' for Reddit dataset. While controlling attribute `sad', we only see an improvement relative to the base DialoGPTm model. The summary of this experiment is presented in the Table ~\ref{tab:human-evaluation-control}.

\begin{table}[]
\centering
\scriptsize
\begin{tabular}{|c|c|c|c|c|c|}
\hline
\textbf{Model} & \textbf{Polite} ($\uparrow$) & \textbf{Joy} ($\uparrow$) & \textbf{Anger} ($\uparrow$)& \textbf{Sad} ($\uparrow$)& \textbf{Fear} ($\uparrow$)\\\hline

\multicolumn{6}{|c|}{\textbf{CONAN}}                    \\ \hline
GPS         & 0.50 & 1.30 & 2.50 & \textbf{1.00} & 0.00 \\
DGPTm    & 0.59 & \textbf{2.50} & 3.00 & 0.75 & 0.75 \\
DGPTm-c & \textbf{2.00} & 1.00 & \textbf{4.00} & \textbf{1.00} & \textbf{2.00} \\ \hline
\multicolumn{6}{|c|}{\textbf{Reddit}}                   \\ \hline
GPS         & 1.83 & 0.93 & 1.50 & 0.33 & 0.36 \\
DGPTm    & 2.66 & 2.50 & 1.50 & 0.66 & \textbf{1.33} \\
DGPTm-c & \textbf{3.50} & \textbf{3.33} & \textbf{2.00} & \textbf{2.00} & 1.25 \\ \hline
\multicolumn{6}{|c|}{\textbf{Gab}}                      \\ \hline
GPS         & 1.56 & 1.28 & 0.81 & 0.4  & 0.17 \\
DGPTm    & 2.17 & 2.50 & 1.66 & 1.11 & 0.89 \\
DGPTm-c & \textbf{3.21} & \textbf{2.92} & \textbf{1.90} &\textbf{ 2.03} & \textbf{1.00}\\ \hline
\end{tabular}
\caption{\scriptsize{Average human judgement scores (scale 0-5) for each of the models -- GPS, DialoGPTm and controlled DialoGPTm (DGPTm). Each column represents the attribute that DialoGPTm-c (DGPTm-c) is controlled for. For all the metrics, higher is better and \textbf{bold} indicates best scores.} 
}
\label{tab:human-evaluation-control}
\end{table}

\section{Conclusion and future work}
Our research aims to add  controllable parameters to counterspeech generation setup which can help the moderators to tune the counterspeech toward a particular strategy. Our controllable \textsc{GeDi} models for six different attribute shows significant improvement in the attribute scores over the baselines. We also try to control the generation using multi-attribute and find that the attribute scores can increase further if suitable attributes are mixed together. 

In the future, we plan to add other attributes like `hope'~\cite{palakodety2020hope} to the controllable generation pipeline. 
Finally, we would aim to build a counterspeech suggestion tool around this setup and allow counter speakers (NGO operators/moderators) to control the generation output as per their query attribute(s).

\section*{Ethical impact} Hate speech is a complex phenomenon. While the language generation methods are better than before, it is still very far from generating coherent and meaningful replies~\cite{10.1145/3442188.3445922}. Hence, we advocate against deployment of fully automatic pipelines for countering hate speech~\cite{delosRiscos2021}. Based on the current progress, in this pipeline, an active participation of the counter speakers is required to generate relevant counterspeech. 
This automation, in turn, has the potential to reduce the mental toll of the counter speakers, at least partially.
\bibliographystyle{named}
\small{\bibliography{ijcai_camera_ready}}

\appendix

\section{Ablation study}

In order to further understand the influence of each attribute, we perform an ablation study on the multi-attribute setups. For each setup, we remove an attribute and generate the sentences for the other two attributes. Finally, we measure the score for that removed attribute itself. We report the summary of the results in Table \ref{tab:multi-attribute-ablation-conan} for CONAN, Table \ref{tab:multi-attribute-ablation-reddit} for Reddit and  Table \ref{tab:multi-attribute-ablation-gab} for Gab dataset. When the detox attribute is removed, we do not see much change in the detoxification score (around 1-2\% drop) across all datasets. On the other hand, removal of the politeness attribute decreases the scores massively. We observe an average of 12\%, 15\% and 14\% drops across CONAN, Reddit and Gab datasets respectively.

Among the emotions, when the `joy' attribute is removed we observe a huge reduction in the attribute score for the CONAN dataset (24\%), while for other datasets the drop remains below 10\%. Most significant change in the emotion score takes place when removing `anger' and `sadness' attributes where the average reduction remains around 40-60\% across all the datasets. Finally, when removing `fear' attribute, we only see a change for CONAN dataset (83\%) but other scores remain almost the same.
\begin{table}[!htpb]
\centering
\footnotesize{
\begin{tabular}{|c|c|c|c|}
\hline
\textbf{Attributes}          & \textbf{Detox}              & \textbf{Polite}           & \textbf{Emotion} \\ \hline
Joy(J)+Polite   &    0.73             &  --               & --        \\ 
Joy+Detox    &    --                &  3.44             & --        \\ 
Polite+Detox    &    --                &  --               & 0.37 (J)        \\ \hline
Anger(A)+Polite &    0.68            &  --               & --        \\ 
Anger+Detox  &     --             &  2.79             & --        \\ 
Polite+Detox    &     --             &  --               & 0.05 (A)       \\ \hline
Sad(S)+Polite   &     0.69             & --               & --        \\ \
Sad+Detox   &     --             &  3.20               & --        \\ 
Polite+Detox   &     --             &  --               & 0.03 (S)        \\ \hline
Fear(F)+Polite  &     0.70           & --               & --        \\ 
Fear+Detox  &      --             &  3.30            & --        \\
Polite+Detox  &       --             &  --               & 0.01 (F)        \\ \hline

\end{tabular}
}
\caption{\footnotesize{Results of the ablation study for $DialoGPT_{medium}$ model trained on CONAN dataset. In each of these setups, we remove one of the attribute and re-estimate that attribute's score. The last column -- \textit{emotion} represents the score of the emotion that is being controlled for that instance.}}
\label{tab:multi-attribute-ablation-conan}
\end{table}

\begin{table}[!htpb]
\centering
\footnotesize{
\begin{tabular}{|c|c|c|c|}
\hline
\textbf{Attributes}          & \textbf{Detox}              & \textbf{Polite}           & \textbf{Emotion} \\ \hline
Joy(J)+Polite    &  0.87               & --                & --        \\ 
Joy+Detox    &  --                  & 5.12              & --        \\
Polite+Detox    &  --               & --                & 0.76 (J)       \\ \hline
Anger(A)+Polite  &  0.82               & --                & --    \\ 
Anger+Detox  &   --                 & 3.46              & --    \\ 
Polite+Detox  &  --                 & --                & 0.09 (A)    \\ \hline
Sad(S)+Polite    &  0.84               & --                & --       \\ 
Sad+Detox     &   --                & 3.96               & --       \\ 
Polite+Detox  &  --                 & --                & 0.05 (S)       \\ \hline
Fear(F)+Polite   &  0.86               & --                &        \\ 
Fear+Detox   &   --                 & 3.34              & --       \\ 
Polite+Detox   &  --                & --                & 0.01 (F)       \\ \hline

\end{tabular}
}
\caption{\footnotesize{Results of the ablation study for $DialoGPT_{medium}$ model trained on Reddit dataset. In each of these setups, we remove one of the attribute and re-estimate that attribute's score. The last column -- \textit{emotion} represents the score of the emotion that is being controlled for that instance.}}
\label{tab:multi-attribute-ablation-reddit}
\end{table}

\begin{table}[!htpb]
\centering
\footnotesize
\begin{tabular}{|c|c|c|c|}
\hline
\textbf{Attributes}          & \textbf{Detox}              & \textbf{Polite}           & \textbf{Emotion} \\ \hline
Joy(J)+Polite    & 0.85                & --                & -- \\ 
Joy+Detox    &  --                  & 5.09              & -- \\ 
Polite+Detox    & --                & --                & 0.82 (J) \\ \hline

Anger(A)+Polite  & 0.80                & --                & --  \\ 
Anger+Detox     & --                & 3.41              & --  \\ 
Polite+Detox     & --                & --               & 0.08 (A)  \\ \hline

Sad(S)+Polite    & 0.82                & --                & -- \\ 
Sad+Detox     &  --                & 4.19                & -- \\ 
Polite+Detox  & --                & ---                & 0.04 (S) \\ \hline

Fear(F)+Polite   & 0.85             & --                & --\\ 
Fear+Detox   & --               & 4.69                & --\\ 
Polite+Detox   & --                & --                & 0.00 (F)\\ \hline

\end{tabular}
\caption{\footnotesize{Results of the ablation study for $DialoGPT_{medium}$ model trained on Gab dataset. In each of these setups, we remove one of the attribute and re-estimate that attribute's score. The last column -- \textit{emotion} represents the score of the emotion that is being controlled for that instance.}}
\label{tab:multi-attribute-ablation-gab}
\end{table}
\section{Metrics}
The diversity~\cite{wang2018sentigan} of the given set of generated sentences $s$ is defined in equation  \ref{eq:7}. $\psi$ is the Jaccard similarity function.
\begin{equation}
\footnotesize
\label{eq:7}
diversity(s) =(1/|s|)*\sum_{i}1-max((\psi(s_{i},s_{j}))_{j=1}^{j=|s|,j!=i}
\end{equation}

Finally, we measure the novelty of the generated outputs to understand if the outputs are directly copied from the training dataset or not. We calculate the novelty~\cite{wang2018sentigan} using equation \ref{eq:8} where $c$ is the sentence set of training corpus and $\psi$ is the Jaccard similarity function. 

\begin{equation}
\footnotesize
    \label{eq:8}
    novelty(s) =(1/|s|)*\sum_{i}1-max((\psi(s_{i},c_{j}))_{j=1}^{j=|c|}
\end{equation}

\section{Other hyperparameters}
For the generation module, we fix the maximum generation length at 100 tokens due to resource constraints. \textit{No repeat ngram size} was kept at 5, \textit{repetition penalty} was set at 3.5 and \textit{temperature} for generating output was 1.2. These steps were taken to further enhance the diversity of the samples generated.We also use nucleus sampling as a decoding strategy~\cite{holtzman2019curious} with $k=0.92$ and $p=100$ to increase the diversity of the outputs.

\section{Limitations}

We rely on the public datasets available for our work. Hence, the generated counterspeech is to some extent based on the properties of the dataset. For example, the CONAN dataset is primarily focused on countering Islamophobic speech. Second, this framework may not directly generalise to more complex type of counterspeech like contradiction or sarcasm.

\section{Examples of controlled generation}
We present examples from controlled counterspeech generation corresponding to few hate speech. \ref{tab:table-example2} and \ref{tab:table-example3} show the generation based on single attribute control, while Tables \ref{tab:table-example4} and \ref{tab:table-example5} show the generated outputs for multi-attribute control.

\begin{table}[!htpb]
\centering
\footnotesize{
\begin{tabular}{p{2cm}p{5cm}}\hline
Hate speech              &  \textit{Never break bread with a Quranimal. You risk catching HIV and/or Ebola.}.\\\hline
Counter speech (fear)    &  \textit{How can you expect us to live in peaceful culture if we are afraid of everything that might come after? how can tolerantism do anything but bring fear? also it means focus on your chosen faith not get influenced by outsiders.}\\\hline
Counterspeech (sadness) &  \textit{how is having people over spreading it such an issue, could you elaborate on this hatred of the quran for how these disgusting disease spread from us to their home countries. please do! its just so sad indeed. not everyone should tolerate hate like this.}\\\hline
Counterspeech (anger) &  \textit{So not true. Islam can lead the way to salvation, but no one believes that. We should respect diversity! And you know, curing other peoples diseases? That seems very selfish... we should eradicate disease!}\\\hline

\end{tabular}}
\caption{\footnotesize{This table presents an example hate speech and the counterspeech generation using different control attributes which is marked in the (bracket).}}
\label{tab:table-example2}
\end{table}

\begin{table}[!htpb]
\centering
\footnotesize{
\begin{tabular}{p{2cm}p{5cm}}\hline
Hate speech              &  \textit{wats wrong with being a commie jew? don't you want that high-iq baby inside your nazi womb?}.\\\hline
Counterspeech (politeness)    &  \textit{there is no need to degrade an entire religion based on the actions of a few people here, please moderate your speech outside of this space and think about how using such language could be hurtful or even unfunny. thank you!  happy  new year!}\\\hline
Counter speech (joy) &  \textit{its inappropriate to use this word when referring specifically to muslims and can be viewed as racist. please do not use these terms or consider them hurtful. thank you for using better language in the future.}\\\hline
Counterspeech (sad) &  \textit{making fun of nationality, religion and sexual preference is not kosher in todays world. try rereading them both times just in case they are harmful for you or anyone! first time using words used here makes those hurt even more}\\\hline
\end{tabular}}
\caption{\footnotesize{This table presents an example hate speech and the counter speech generation using different control attributes which is marked in the (bracket).}}
\label{tab:table-example3}
\end{table}

\begin{table}[!htpb]
\centering
\footnotesize{
\begin{tabular}{p{2cm}p{5cm}}\hline
Hate speech              &  \textit{A black person robbed my friend so fuck you black pieces of shit. One day we will lynch all of you cuz you had it coming.
}.\\\hline
Counterspeech (joy + polite)    &  \textit{to use hateful language only shows how intolerant and bigoted you are, in the hopes that you can be more accepting with your experiences. remember to use better language yourself in the future. thank you for your understanding.}\\\hline
Counter speech (anger + polite) &  \textit{wow. what a hate filled hateful word that should not be used when discussing the color of peoples skin! several such things can contribute to productive discussion, and any further comments could easily be summarized using such words}\\\hline
\end{tabular}}
\caption{\footnotesize{This table presents an example hate speech and the counter speech generation using multi-attributes control, which are marked in the (bracket).}}
\label{tab:table-example4}
\end{table}

\begin{table}[!htpb]
\centering
\footnotesize{
\begin{tabular}{p{2cm}p{5cm}}\hline
Hate speech              &  \textit{Remember that time a bunch of people kidnapped a retard and tortured him live?}.\\\hline
Counter speech (joy + polite)    &  \textit{it only happened once but in this instance its not ok to use pejoratively based on medical conditions if you like supporting your opinions further so please feel free to vote out any language that suggests 
it may be used. thanks}\\\hline
Counter speech (anger + toxic) &  \textit{wow, how racist these comments are. it actually incredibly disrespectful to generalize people based on their race. we see plenty of genetic variation among people, not just racial variance. 
let us not thank anyone for missing details
}\\\hline
Counter speech (sadness + polite)    &  \textit{i have no idea why you are even posting this. reading through such posts hurt 
deep down inside with fear by where your women will cry out from these comments when compared to actual victims experiences across history s repulsive has good won t it now would yield illiberal, whatever issues acusot or moronic could look like. alsoplease refrain from using mental illness as an insult ( over here ) thanks for the heads up}\\\hline
\end{tabular}}
\caption{\footnotesize{This table presents an example hate speech and the counterspeech generation using multi-attributes control, which are marked in the (bracket).}}
\label{tab:table-example5}
\end{table}

\section{System and code requirements}
For all the experiments in this paper, we used 48-core Xeon processor Linux based system with 126 GB RAM. For training  the  neural networks  we  used 2 NVIDIA P100 GPUs having 16 GB each with CUDA version 10.1. We primarily based our system on Python libraries. Among the neural networks we used Huggingface's transformers library\footnote{https://huggingface.co/} for GPT-2 based models with PyTorch as backend in general. All the libraries used in this research are pip installable. Further we also resort to the code which controls the generation using \textsc{GeDi} models and the code which trains the \textsc{GeDi} models from the authors' git repository\footnote{https://github.com/salesforce/GeDi}.

\section{Human judgement details}
\begin{figure}
    \centering
    \includegraphics[width=\textwidth]{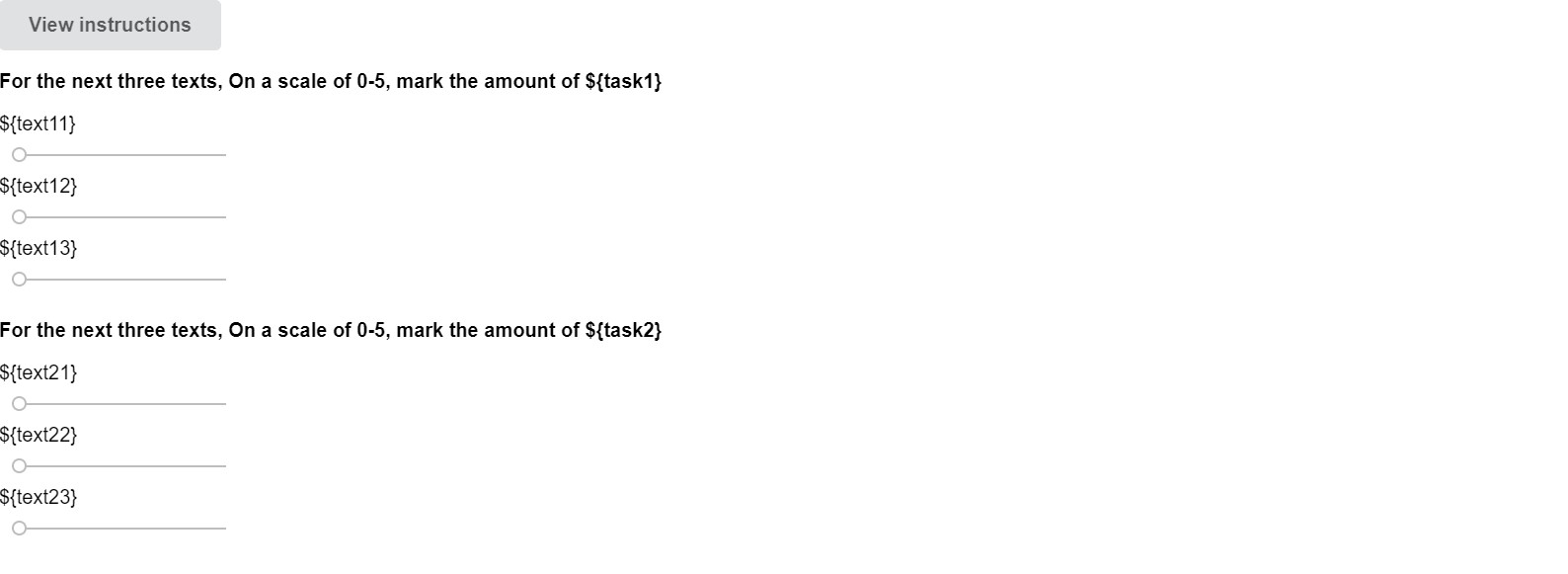}
    \caption{The interface design for the Amazon Mechanical Turk platform.}
    \label{fig:interface}
\end{figure}

The annotators include 2 PhD and 3 BTech students. We consider the definitions and use several examples from the relevant attribute datasets to provide examples to the annotators to help them mark the presence of that attribute in the presented counterspeech. The final interface is shown in Figure \ref{fig:interface}. We use  Amazon Mechanical Turk (AMT) sandbox\footnote{https://requestersandbox.mturk.com/create/projects} environment, where the annotators login using their account and annotate the examples.

\end{document}